\documentclass[lettersize,journal]{IEEEtran}
\usepackage{amsmath,amsfonts}
\usepackage{algorithmic}
\usepackage{algorithm}
\usepackage{array}
\usepackage[caption=false,font=normalsize,labelfont=sf,textfont=sf]{subfig}
\usepackage{textcomp}
\usepackage{stfloats}
\usepackage{booktabs}
\usepackage{url}
\usepackage{verbatim}
\usepackage{graphicx}
\usepackage{cite}
\usepackage{times}
\usepackage{mathptmx}
\usepackage{newtxtext,newtxmath}
\usepackage[table]{xcolor}   
\begin{document}

\title{CoLA-Flow Policy: Temporally Coherent Imitation Learning via Continuous Latent Action Flow Matching for Robotic Manipulation}

\markboth{IEEE ROBOTICS AND AUTOMATION LETTERS. PREPRINT VERSION. ACCEPTED JUNE, 2026}{Wu et al.: CoLA-Flow Policy for Robotic Manipulation}
\def\leftmark{IEEE ROBOTICS AND AUTOMATION LETTERS. PREPRINT VERSION. ACCEPTED JUNE, 2026}
\def\rightmark{Wu et al.: CoLA-Flow Policy for Robotic Manipulation}
\makeatletter
\def\ps@headings{%
  \def\@oddhead{\ifodd\c@page\hbox{}\@IEEEheaderstyle\rightmark\hfil\thepage\else\@IEEEheaderstyle\thepage\hfil\leftmark\hbox{}\fi}\relax
  \def\@evenhead{\@IEEEheaderstyle\thepage\hfil\leftmark\hbox{}}\relax
  \let\@oddfoot\@empty
  \let\@evenfoot\@empty}
\def\ps@IEEEtitlepagestyle{%
  \def\@oddhead{\hbox{}\@IEEEheaderstyle\leftmark\hfil\thepage}\relax
  \def\@evenhead{\@IEEEheaderstyle\thepage\hfil\leftmark\hbox{}}\relax
  \let\@oddfoot\@empty
  \let\@evenfoot\@empty}
\makeatother

\author{Songwei Wu, Zhiduo Jiang, Wandong Sun, Guanghu Xie, Yuteng Xie, Rui Zhao, Yang Liu, and Hong Liu%
\thanks{Manuscript received: February 5, 2026; Revised: April 12, 2026; \mbox{Accepted: June 13, 2026.}

This paper was recommended for publication by Editor Wei Pan upon evaluation of the Associate Editor and Reviewers comments.
This work was supported by the National Natural Science Foundation of China under the Basic Science Center Program for ``Space Robot Intelligent Manipulation'' (Grant No. T2388101) and the Natural Science Foundation of Heilongjiang Province for Excellent Young Scholars (Grant No. YQ2024E018). (Corresponding author: Yang Liu.)

Songwei Wu, Zhiduo Jiang, Wandong Sun, Guanghu Xie, Yang Liu, and Hong Liu are with the State Key Laboratory of Robotics and System, Harbin Institute of Technology, Harbin 150001, China (e-mail: \{24S008057, 24S108210, 24B908020, 23B308003\}@stu.hit.edu.cn; \{liuyanghit, hong.liu\}@hit.edu.cn).
Yuteng Xie is with The University of Sydney, Sydney, NSW 2006, Australia (e-mail: yxie0956@uni.sydney.edu.au).
Rui Zhao is with Honor Device Co., Ltd., Shenzhen, China (e-mail: rui.zhao.ml@gmail.com).

Digital Object Identifier (DOI): see top of this page.}
}


\maketitle
\thispagestyle{IEEEtitlepagestyle}
\pagestyle{headings}

\begin{abstract}
Learning long-horizon robotic manipulation requires jointly achieving expressive behavior modeling,
real-time inference, and stable execution, which remains challenging for existing generative policies.
Diffusion-based approaches offer strong modeling capacity but incur high inference latency,
while flow matching enables fast, near-single-step generation yet often suffers from unstable execution when operating directly in the raw action space.
We propose \textbf{Continuous Latent Action Flow Policy (CoLA-Flow Policy)}, a trajectory-level imitation learning framework
that performs flow matching in a continuous latent action space.
By encoding action sequences into temporally coherent latent trajectories and learning an explicit latent-space flow,
CoLA-Flow Policy decouples global motion structure from low-level control noise, enabling smooth and reliable long-horizon execution.
The framework further integrates geometry-aware point cloud conditioning and execution-time multimodal modulation, using visual cues as a representative modality to enhance real-world robustness.
Experiments in simulation and on real robots show that CoLA-Flow Policy achieves near-single-step inference,
improves trajectory smoothness by up to \textbf{93.7\%} and task success by up to \textbf{25 percentage points} over raw action-space flow baselines,
while remaining significantly faster than diffusion-based policies.
\end{abstract}


\section{INTRODUCTION}

Diffusion-based policies (DP)~\cite{DP_INI} have demonstrated strong performance in robotic visuomotor imitation learning,
especially for complex manipulation tasks in simulation and real-world environments.
By modeling multimodal action distributions conditioned on images~\cite{DP_INI,ni2025vo} or point clouds~\cite{DP3,dexcap,HDP},
these methods enable expressive behavior modeling across diverse task variations.

Flow matching~\cite{FM,chen2023flow} has recently emerged as an efficient alternative for robotic control~\cite{FP,RFM,pi05}.
By formulating action generation as an ODE-based transport process, flow matching enables near-single-step inference,
making it promising for real-time imitation learning.

However, deploying generative policies on physical robots remains challenging:
diffusion policies incur high inference latency from iterative denoising,
while flow-based policies may amplify modeling noise over long horizons,
causing jittery or unstable trajectories in multi-stage tasks.
Thus, expressive behavior modeling, fast inference, and stable execution remain difficult to achieve jointly.

Recent works attempt to alleviate these issues through efficient point cloud encoders~\cite{DP3,iDP3},
hierarchical Fast--Slow control~\cite{RDP}, or consistency constraints~\cite{FP}.
Nevertheless, our experiments show that the trade-off between inference efficiency and execution stability remains unresolved.

To address this challenge, we propose \textbf{Continuous Latent Action Flow Policy (CoLA-Flow Policy)}, a trajectory-level framework that performs flow matching in a continuous latent action space under geometry-aware 3D scene conditioning.
Unlike discrete latent action approaches such as VQ-VAE variants, our continuous latent trajectories help preserve high-precision control and enable smoother transitions.
By generating whole trajectories in a structured latent space rather than directly in the physical action space, CoLA-Flow Policy enables efficient one-step inference while maintaining temporal coherence for long-horizon manipulation.

CoLA-Flow Policy decouples global motion structure from low-level control noise:
by operating on motion-level latent representations, the learned flow captures smoother trajectory evolution and reduces the amplification of modeling errors during generation.
Environment-dependent decoding further incorporates multimodal cues, such as wrist-camera feedback at execution time, without altering the generative process, improving smoothness and robustness in long-horizon manipulation.

Our main contributions are summarized as follows:
\begin{itemize}
    \item We propose a trajectory-level latent flow matching framework and experimentally demonstrate its ability to balance inference efficiency with temporal consistency in complex, long-horizon manipulation.
    \item We show that our continuous latent action space suppresses trajectory jitter and supports execution-time multimodal modulation, improving execution stability and adaptability.
    \item We validate the method through simulation, real-robot tasks, and additional long-horizon, modality, and perturbation ablations against discrete VQ-VAE latent and no-latent variants.
\end{itemize}

\begin{figure*}[t]
    \centering
    \includegraphics[width=0.9\textwidth,trim=20pt 20pt 20pt 20pt,
        clip]{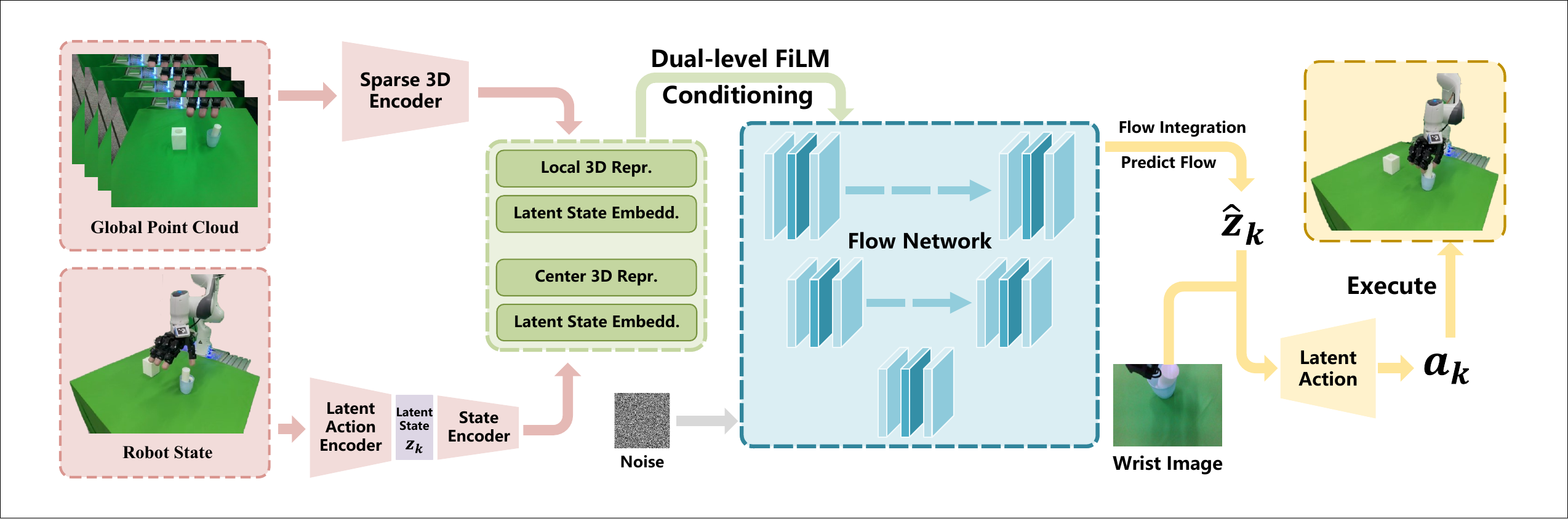}
    \caption{Overall architecture of CoLA-Flow Policy: point cloud observations are encoded into geometry-aware features, latent-space flow matching generates coherent latent trajectories, and visual conditioning modulates action decoding.}
    \label{fig:overall_arch}
\end{figure*}

\section{RELATED WORKS}

\subsection{Diffusion and Flow-Based Policies for Robotics}

Diffusion-based policies~\cite{Diffusion,DP_INI,DP3,iDP3,HDP,ni2025vo} formulate visuomotor control as conditional trajectory generation and perform well in complex manipulation tasks.
However, iterative denoising introduces high latency and sensitivity to execution noise; even with DDIM acceleration~\cite{DDIM} or hierarchical designs such as RDP~\cite{RDP}, multi-step generation remains a bottleneck.

Flow-based generative models~\cite{FM,chen2023flow} offer a more direct alternative by learning ODE-based probability paths.
Flow matching enables faster inference by regressing velocity fields, and recent consistency-based variants~\cite{FP,RFM} have extended this idea to policy learning, alongside efficient policy models~\cite{pi05}.
Nevertheless, directly applying flow-based policies in the raw action space can still cause trajectory oscillations under high-dimensional actions and complex visual inputs.

Our work builds on consistency flow matching and addresses this limitation through trajectory-level latent modeling and geometry-aware conditioning.

\subsection{Visual Imitation Learning with 3D Representations}

Early visual imitation learning methods rely primarily on 2D image observations~\cite{DP_INI,florence2022implicit},
processed by convolutional networks~\cite{CNN} or vision transformers~\cite{ViT}.
Recent 3D-based approaches~\cite{DP3,iDP3,dexcap} incorporate depth or point clouds for geometric reasoning,
while 3D-aware VLAs introduce spatial cues for language-conditioned manipulation.
For example, MolmoAct~\cite{moact} uses depth perception tokens and visual reasoning traces,
while OG-VLA~\cite{ogvla} renders RGB-D point clouds into canonical orthographic views for keyframe prediction.

However, these representations are often coupled with reasoning-token generation, keyframe prediction, or planning-based execution.
Such formulations~\cite{dexgraspnet,dexgraspnet2,ogvla} add test-time decoding or optimization and scale poorly to high-dimensional continuous control.
Meanwhile, efficient point cloud encoding remains challenging: expressive encoders can be computationally expensive~\cite{dexcap},
whereas simplified ones~\cite{DP3,iDP3} may lose fine-grained geometric sensitivity.

In contrast, our environment perception unit directly encodes structured point-cloud geometry for latent action generation, providing scene-level spatial awareness without depth-token reasoning, orthographic rendering, or test-time keyframe planning, thereby supporting real-time continuous trajectory prediction.

\subsection{Latent Action Learning}
Latent action learning introduces compact intermediate representations for robot control, improving temporal abstraction and action consistency~\cite{RDP,clam,world}.
One strategy maps continuous action trajectories into discrete latent tokens through VQ-VAE-style quantization~\cite{vqvae} or action tokenization used in large-scale policy pre-training~\cite{pi05,KI}.
However, discretizing continuous robot actions can discard fine-grained motion information, limiting smooth and precise control.
RDP~\cite{RDP} follows this direction by using a VQ-VAE-style discrete latent action representation together with a diffusion-style slow--fast reactive generation framework.
In contrast, CoLA-Flow learns a GRU-based continuous latent trajectory representation and performs flow matching directly in this continuous latent action space; wrist-camera feedback is introduced only as decoder-side execution-time modulation, without changing the latent flow dynamics.

Continuous latent actions have also been explored: 
CLAM~\cite{clam} infers latent actions from action-free demonstrations via inverse and forward dynamics models, while DreamZero~\cite{world} predicts future videos and action chunks with an autoregressive diffusion transformer.
These methods mainly target action-free relabeling or video-based world modeling and often involve additional decoding or iterative generation.

In contrast, CoLA-Flow models executable actions as continuous latent trajectories within a consistency flow matching framework, avoiding discretization-induced information loss and iterative sampling latency while enabling temporally consistent action generation conditioned on structured 3D geometry.

\section{METHODS}

We propose CoLA-Flow Policy, a modular imitation learning framework for long-horizon robotic manipulation.
As illustrated in Fig.~\ref{fig:overall_arch}, the method first learns a temporally coherent latent action representation,
which serves as a compact trajectory-level abstraction.
A generative model then operates in this latent space to produce long-horizon latent trajectories under geometric scene conditioning,
and the resulting latent actions are finally decoded into executable control commands with execution-time sensory modulation.

\subsection{Coherent Latent Action Representation}
\label{sec:latent_action_rep}

\begin{figure}[t]
    \centering
    \includegraphics[width=0.45\textwidth,trim=20pt 20pt 20pt 20pt,clip]{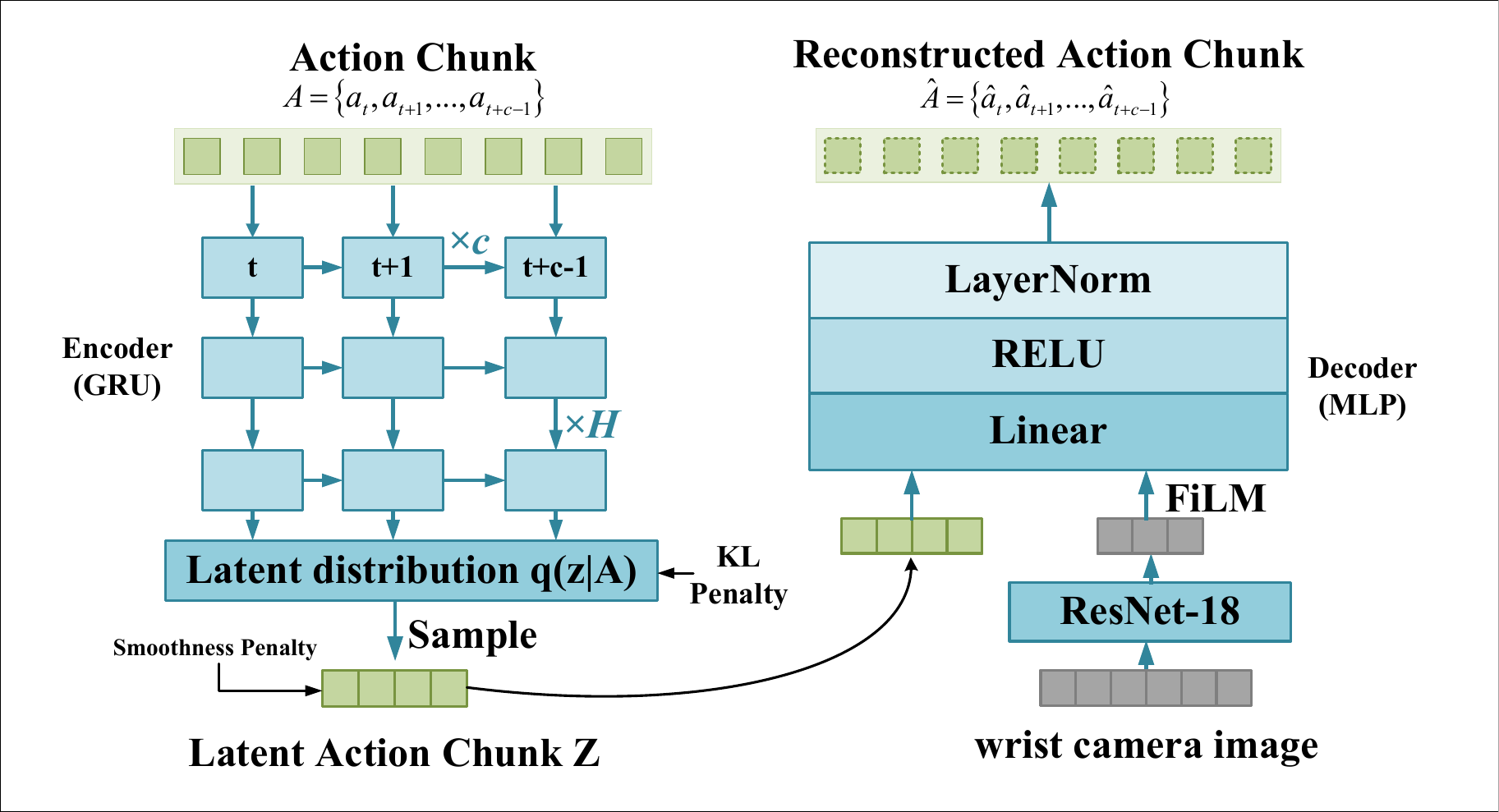}
    \caption{Trajectory-level latent action representation with recurrent encoding and conditional decoding.
    A GRU-based encoder introduces a temporal inductive bias that promotes coherent latent trajectories,
    while an MLP decoder modulated by wrist-camera features via FiLM enables visually adaptive action execution.}
    \label{fig:latent_arch}
\end{figure}

Long-horizon robotic manipulation requires action trajectories that are both expressive and execution-stable.
Rather than modeling single-step actions, we represent actions as temporally extended chunks, allowing the latent space to capture short-horizon motion patterns and local temporal coherence.
Since applying flow-based generative models directly in the raw action space can amplify numerical errors and high-frequency noise during trajectory integration, we introduce a \emph{trajectory-level latent action modeling framework}.
It encodes action chunks into a continuous latent trajectory space and decodes generated latent trajectories into executable controls, as illustrated in Fig.~\ref{fig:latent_arch}.

\paragraph{Temporally Continuous Latent Action Modeling}
Given an action segment $\mathbf{A}_{t:t+L-1} \in \mathbb{R}^{L \times d_a}$, we partition it into $K$ local segments
$\{\mathbf{A}^{(1)}, \ldots, \mathbf{A}^{(K)}\}$, each spanning $c$ time steps.
Each segment is first embedded by a lightweight temporal convolution and then processed by a GRU encoder:
\begin{equation}
    \mathbf{h}_k = \mathrm{GRU}(\mathbf{x}_k, \mathbf{h}_{k-1})
\end{equation}
where $\mathbf{x}_k$ denotes the embedding of the $k$-th local action segment.

The latent code $\mathbf{z}_k$ is inferred from $\mathbf{h}_k$, forming a history-dependent latent trajectory across consecutive action chunks.
By conditioning each latent code on previous segments, the recurrent encoder introduces a temporal inductive bias for smooth generation.
This shapes the latent action space into a coherent trajectory manifold, which is beneficial for stable ODE-based flow matching.
In contrast, non-recurrent per-segment encoders break temporal continuity and empirically lead to increased trajectory jitter.

\paragraph{Variational Regularization}
To further regularize the latent space, we adopt a variational formulation~\cite{vae}.
A KL divergence term between the posterior and a standard normal prior encourages compactness and continuity of the latent distribution,
improving robustness to noisy demonstrations.
This regularization operates at the trajectory level, making it well suited for long-horizon manipulation.

\paragraph{Multimodal-Conditioned Action Decoding}
The decoder incorporates execution-time sensory information without interfering with latent trajectory generation.
Although this work uses wrist-mounted camera observations, as shown in Fig.~\ref{fig:latent_arch}, the design is not restricted to wrist-view images.
When available, other task-relevant modalities, such as external vision, tactile feedback, force/torque signals, or proprioception, can also be integrated as auxiliary inputs through feature-wise linear modulation (FiLM)~\cite{film}.
In our implementation, wrist-view images are encoded by a lightweight visual encoder based on a pretrained ResNet-18~\cite{resnet}, and the resulting features are injected into the decoder via FiLM layers.

Given a latent action code $\mathbf{z}_k$, the decoder reconstructs the corresponding action segment as
\begin{equation}
\hat{\mathbf{A}}^{(k)} = f_\theta(\mathbf{z}_k \mid \mathbf{v}_t)
\end{equation}
where $\mathbf{v}_t$ denotes the encoded wrist-view visual feature in our implementation.
By confining sensory conditioning to the decoding stage, the framework cleanly separates latent trajectory planning
from execution-time adaptation while preserving temporal coherence.

\paragraph{Training Objective and Protocol}
The latent action encoder--decoder is first trained using a variational objective with reconstruction,
KL regularization, and a lightweight smoothness constraint to learn a temporally coherent latent space.
The learned representation is then fixed and used as a plug-in interface,
upon which a downstream generative model is trained to generate long-horizon latent trajectories.

\subsection{Latent-Space Action Generation via Flow Matching}

Given a latent action trajectory $\mathbf{Z}=\{\mathbf{z}_1, \ldots, \mathbf{z}_K\}$ obtained from the trajectory-level encoder,
we generate long-horizon manipulation behaviors by modeling their evolution in continuous time using flow matching~\cite{FM}.
Unlike diffusion-based policies~\cite{DP_INI} that rely on iterative denoising, flow matching enables efficient near-single-step generation.
However, when applied directly in the raw action space, flow-based models are highly sensitive to local inconsistencies and noise.
We therefore perform flow matching exclusively in the temporally coherent latent action space, which is explicitly designed to be smooth and temporally coherent
(Sec.~\ref{sec:latent_action_rep}).

\paragraph{Latent-Space Flow Dynamics}
Let $\mathbf{z}$ denote a latent action code sampled from the target latent distribution and $\tilde{\mathbf{z}}$ a corresponding sample from a simple base distribution.
We learn a time-dependent vector field $\boldsymbol{\nu}_\theta(\tau, \mathbf{z})$ that defines a continuous probability path from the source to the target distribution,
governed by the ordinary differential equation
\begin{equation}
    \frac{d \xi_{\mathbf{z}}(\tau)}{d\tau} =
    \boldsymbol{\nu}_\theta(\tau, \xi_{\mathbf{z}}(\tau)), \quad
    \xi_{\mathbf{z}}(0) = \tilde{\mathbf{z}}
\end{equation}
Operating on motion-level latent representations rather than raw control signals substantially improves numerical stability,
as the smooth geometry of the latent space suppresses the amplification of high-frequency variations during trajectory generation.

\paragraph{Consistency Flow Matching in Latent Space}
To efficiently learn the latent vector field across noise levels, we adopt consistency flow matching (CFM)~\cite{cfm} and define the flow function
\begin{equation}
    f_\theta(\tau, \mathbf{z}) =
    \mathbf{z} + (1 - \tau)\, \boldsymbol{\nu}_\theta(\tau, \mathbf{z})
\end{equation}
where $\tau \in [0,1]$ interpolates between the source and target latent distributions.
Here, $\boldsymbol{\nu}_\theta(\tau, \mathbf{z})$ represents a learned velocity field in the latent action space, and the factor $(1 - \tau)$
attenuates update magnitudes as the trajectory approaches the target distribution, improving stability near convergence.
We further apply a time-dependent input normalization
\begin{equation}
    c_{\mathrm{in}}(\tau) = \frac{1}{\sqrt{\tau^2 + (1 - \tau)^2}}
\end{equation}
scaling the latent state before feeding it into the velocity field network: $\boldsymbol{\nu}_\theta\!\left(\tau,\, c_{\mathrm{in}}(\tau)\cdot \mathbf{z}\right)$.
This normalization balances the scales of latent states across noise levels and prevents overfitting to specific time regions.

\paragraph{One-Step Latent Trajectory Generation}
At inference time, latent action codes are sampled from the base distribution and transformed
via a single evaluation of the learned flow:
\begin{equation}
    \hat{\mathbf{z}}_k = f_\theta(0, \tilde{\mathbf{z}}_k)
\end{equation}
As with flow-based generative models, latent trajectories are obtained through a single-step transformation
defined by the learned latent-space flow field, without requiring iterative denoising as in diffusion-based policies~\cite{DP_INI}.
This enables one-shot generation of the entire latent action trajectory, avoiding iterative sampling
and substantially reducing inference latency.

The resulting latent trajectory preserves temporal coherence by construction and is decoded into executable action segments using the
multimodal-conditioned decoder described in Sec.~\ref{sec:latent_action_rep}.
By decoupling trajectory generation from execution-level adaptation, the policy achieves both fast inference and stable, smooth manipulation behavior.

In the following experiments, we compare the proposed CoLA-Flow Policy with two related variants: a discrete VQ-VAE latent baseline and the original Flow Policy. This comparison is designed to evaluate the contribution of the continuous latent action representation and the proposed latent-space flow generation mechanism.

\subsection{Geometry-Aware 3D Scene Conditioning}

Stable execution of latent action trajectories requires accurate scene geometry,
as geometric errors may propagate across decoded action segments over long horizons.
We therefore condition the latent flow policy on a compact 3D scene representation derived from point clouds,
following prior point-cloud conditioned manipulation policies~\cite{DP3}.
Different from the decoder-side sensory modulation in Sec.~\ref{sec:latent_action_rep},
this representation serves as a condition for latent-space action generation.

\paragraph{Point Cloud Encoding}
Scene point clouds are reconstructed from depth images captured by a fixed global camera and cropped within the workspace.
Farthest Point Sampling (FPS)~\cite{pointnet} is then applied to select sparse center points as anchors for local neighborhoods.

\begin{figure}[!t]
    \centering
    \includegraphics[width=0.45\textwidth,height=0.3\textheight,keepaspectratio,trim=20pt 20pt 20pt 20pt,clip]{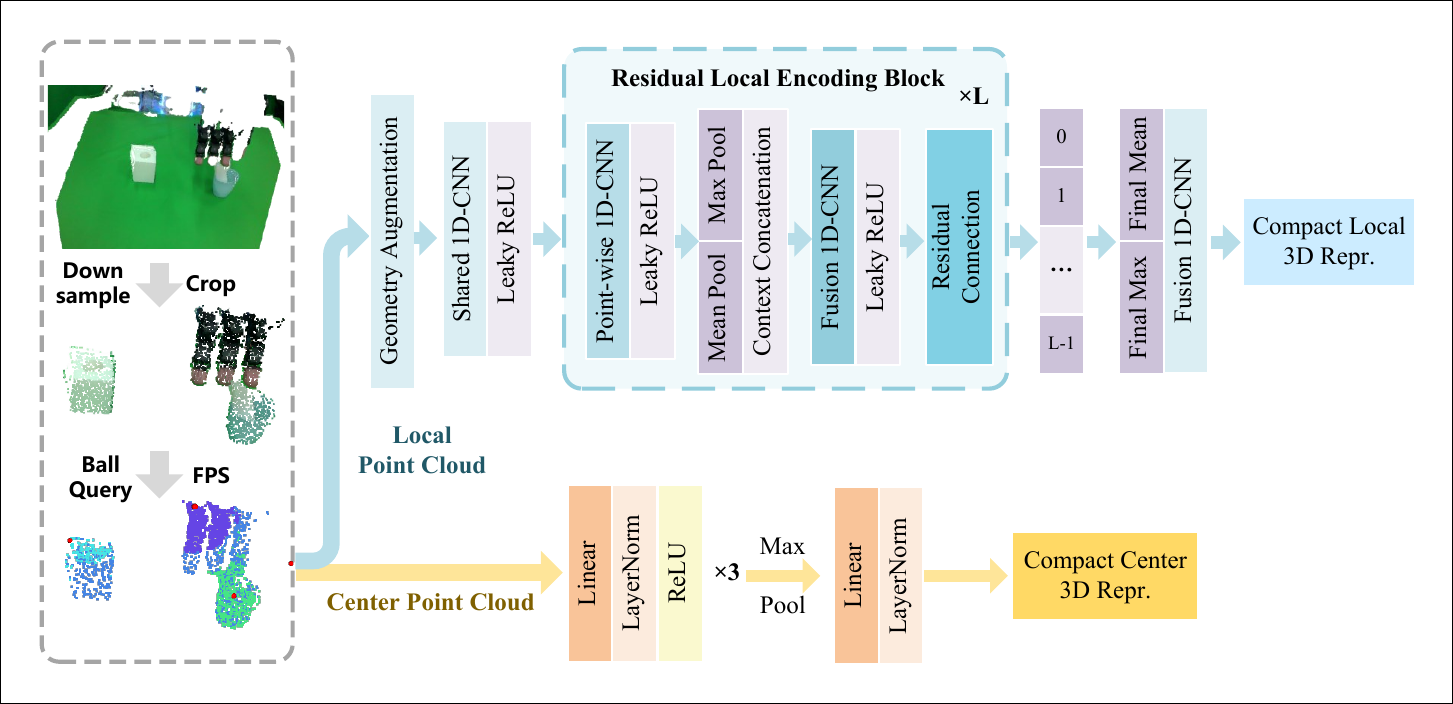}
    \caption{Geometry-aware point cloud encoder.
    Local neighborhoods around farthest-sampled centers capture translation-invariant local geometry via residual convolutions and max--mean pooling, while a lightweight center encoder provides compact global scene context for latent trajectory conditioning.}
    \label{fig:pcd_enc}
\end{figure}

To capture multi-scale geometry, we adopt a dual-branch encoder with a local encoder $f_l$ and a center encoder $f_c$, as shown in Fig.~\ref{fig:pcd_enc}.
The local encoder aggregates relative point offsets within each neighborhood using a residual convolutional backbone with max--mean pooling,
while the center encoder maps sampled center coordinates to a compact global context vector via a lightweight MLP:
\begin{equation}
    \mathbf{f}_l = f_l(\text{pcd}_{\text{local}}), \quad
    \mathbf{f}_c = f_c(\text{pcd}_{\text{center}})
\end{equation}

\paragraph{Hierarchical Geometric Conditioning}
The extracted geometric features are injected into the latent flow network through a two-stage FiLM mechanism.
Given an intermediate activation $\mathbf{h}$, local geometric modulation is applied first,
followed by center-level conditioning:
\begin{align}
    \mathbf{h}' &= \boldsymbol{\gamma}_l \odot \mathbf{h} + \boldsymbol{\beta}_l,
    \quad
    [\boldsymbol{\gamma}_l, \boldsymbol{\beta}_l] = \mathrm{MLP}_l(\mathbf{f}_l) \\
    \mathbf{h}'' &= \boldsymbol{\gamma}_c \odot \mathbf{h}' + \boldsymbol{\beta}_c,
    \quad
    [\boldsymbol{\gamma}_c, \boldsymbol{\beta}_c] = \mathrm{MLP}_c(\mathbf{f}_c)
\end{align}
Applying local modulation before center-level conditioning allows the policy to first adapt to
contact-level constraints and then adjust latent trajectory generation according to the global scene structure.

Overall, this geometry-aware conditioning provides robust spatial context for latent-space action generation,
enabling stable long-horizon manipulation without compromising inference efficiency.

\section{EXPERIMENTS}
\subsection{Evaluation Metrics}
\label{sec:metrics}

We evaluate all methods using three complementary metrics that jointly characterize task performance, real-time efficiency, and execution quality: task success rate, response time, and trajectory smoothness.

\paragraph{Task Success Rate}
Task success rate is defined as the percentage of trials that satisfy task-specific success criteria within the episode horizon.
Success is determined by predefined geometric constraints, including object-pose errors below a threshold and stable grasps or placements maintained for a minimum duration.

\paragraph{Response Time}
Response time measures inference latency from sensory input to action output, reflecting closed-loop responsiveness.

\paragraph{Trajectory Smoothness}
To quantify long-horizon execution stability, we use a trajectory smoothness metric that combines time-domain jerk and frequency-domain oscillation energy.
Given a sequence of actions $\{\mathbf{a}_t\}_{t=1}^{T}$ with control timestep $\Delta t$, the jerk-based term is defined as
\begin{equation}
\mathbf{j}_t =
\frac{\mathbf{a}_{t} - 3\mathbf{a}_{t-1} + 3\mathbf{a}_{t-2} - \mathbf{a}_{t-3}}{(\Delta t)^3},
\quad
\mathcal{S}_{\mathrm{jerk}} =
\frac{1}{T-3} \sum_{t=4}^{T} \| \mathbf{j}_t \|_2^2
\end{equation}
To capture high-frequency oscillations, we further compute the spectral energy ratio from the DFT coefficients $\hat{\mathbf{a}}_f$:
\begin{equation}
\mathcal{S}_{\mathrm{freq}} =
\frac{\sum_{f > f_c} \| \hat{\mathbf{a}}_f \|_2^2}
{\sum_{f} \| \hat{\mathbf{a}}_f \|_2^2},
\quad
\mathcal{S}_{\mathrm{smooth}} =
\alpha \mathcal{S}_{\mathrm{jerk}} + \beta \mathcal{S}_{\mathrm{freq}}
\end{equation}
Here, $f_c$ denotes the cutoff frequency for high-frequency components.
Lower values indicate smoother trajectories with fewer abrupt variations and high-frequency oscillations; after normalizing both terms, we set $\alpha = 0.25$ and $\beta = 0.75$ in all experiments.

\subsection{Experimental Setup and Implementation Details}
\subsubsection{Experimental Setup}
\label{sec:exp_setup}

\paragraph{Simulation Benchmarks}
We evaluate CoLA-Flow Policy on Adroit~\cite{Adroit} and MetaWorld~\cite{meta}, covering 37 MuJoCo-based manipulation tasks~\cite{mujoco}.
Adroit focuses on contact-rich dexterous manipulation with a high-dimensional hand, while MetaWorld includes diverse arm-based tasks of varying difficulty.
For each task, 30 expert demonstrations are collected using well-tuned heuristic policies, and each method is evaluated over 50 rollouts.

\begin{figure*}[!t]
    \centering
    \includegraphics[width=0.95\textwidth,trim=20pt 20pt 20pt 20pt,clip]{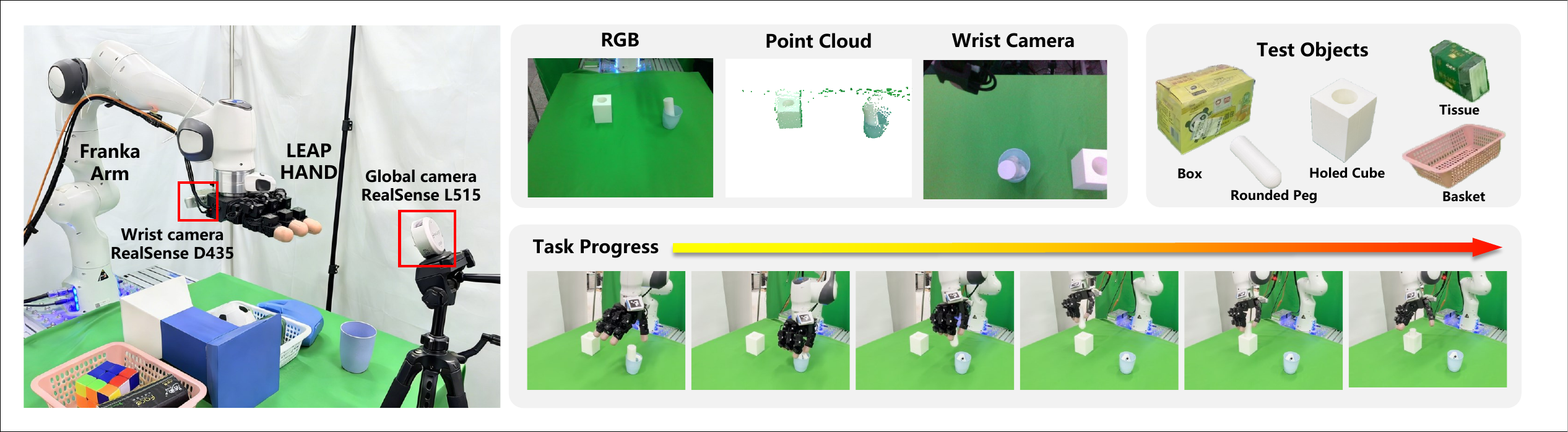}
    \caption{
    Real-world experimental setup and observations.
    Left: Franka Emika Panda robot with a LEAP Hand and the visual sensing setup (global L515 and wrist-mounted D435).
    Right: Example multimodal observations (RGB images, point clouds, wrist views) and task objects used in real-world experiments.
    }
    \label{fig:real_world_scene}
\end{figure*}

\paragraph{Real-World Platform and Tasks}
Real-world experiments use a Franka Emika Panda arm with a LEAP Hand or parallel gripper depending on the task (Fig.~\ref{fig:real_world_scene}, left).
A RealSense L515 global camera provides scene-level RGB-D observations, while a wrist-mounted RealSense D435 camera provides local visual feedback.
Both cameras operate at $640 \times 480$ resolution and 30 fps, and depth observations are converted into point clouds using calibrated intrinsics and extrinsics.
We evaluate four real-world settings: LEAP Hand pick-and-place, gripper pick-and-place, peg-in-hole, and obstacle-avoidance grasping, covering different end-effectors, precise insertion, and clutter-aware grasping; representative objects are shown in Fig.~\ref{fig:real_world_scene} (top right).
For each task, 30 expert demonstrations are collected through human teleoperation.

\paragraph{Baselines and Evaluation Protocol}
We compare CoLA-Flow Policy with Diffusion Policy~\cite{DP_INI}, DP3~\cite{DP3}, iDP3~\cite{iDP3}, Reactive Diffusion Policy (RDP)~\cite{RDP}, and Flow Policy~\cite{FP} under the same simulation protocols.
In real-world experiments, we compare against DP3, RDP, and Flow Policy, adapting RDP to the visual setting by replacing tactile inputs with wrist-camera observations.
Force or pressure sensing is not used due to hardware constraints.
Each real-world task is evaluated over three independent runs with randomized initial conditions, each containing 10 trials.
Success rates are reported as mean $\pm$ standard deviation, and response time is measured as policy inference latency, excluding sensor acquisition and environment stepping.

\subsubsection{Implementation Details}
\label{sec:impl_details}

\begin{table}[t]
  \centering
  \caption{
    Input modalities and action/latent representations across compared policies; wrist images are used only in real-world experiments.
    }
  \label{tab:policy_inputs}
  \renewcommand{\arraystretch}{1.15}
  \footnotesize
  \begin{tabular}{lccc}
    \toprule
    \textbf{Method} &
    \textbf{Point Cloud} &
    \textbf{Wrist Image} &
    \textbf{Action Rep.} \\
    \midrule
    Diffusion Policy~\cite{DP_INI} &
    No (RGB) & No & Raw Action \\
    DP3~\cite{DP3} &
    Yes & No & Raw Action \\
    iDP3~\cite{iDP3} &
    Yes & No & Raw Action \\
    Flow Policy~\cite{FP} &
    Yes & No & Raw Action \\
    RDP~\cite{RDP} &
    Yes & Yes & VQ-VAE latent \\
    \rowcolor{blue!10}
    \textbf{CoLA-Flow Policy} &
    Yes & Yes & GRU latent \\
    \bottomrule
  \end{tabular}
\end{table}

\paragraph{Policy Inputs}
Table~\ref{tab:policy_inputs} summarizes input modalities and action representations.
In simulation, no wrist-camera observations are used, isolating latent action modeling with the same global observations.
In real-world experiments, RDP and CoLA-Flow Policy use wrist-camera observations as additional local visual inputs.

\paragraph{Training and Execution Details}
For all applicable policies, we use a 28-step horizon with 12 observation steps and a 16-step predicted action chunk.
During receding-horizon execution, the first 8 actions are executed before replanning, giving an 8-step overlap between adjacent chunks.
Demonstration trajectories are segmented using the same chunking protocol during training.
We define a task as long-horizon if its execution duration exceeds 100\,s and involves multiple sequential prediction-execution cycles.
RGB observations are cropped to $84 \times 84$, and point clouds are downsampled to 512 points using farthest point sampling.
State and action inputs are normalized to $[-1,1]$, with actions unnormalized before execution.
All models are trained for 150 epochs using AdamW with a learning rate of $1 \times 10^{-4}$, batch size 96, and EMA with a decay rate of 0.95.
Simulation experiments are conducted on an Intel Core i7-14700KF CPU with an NVIDIA RTX 4090D GPU, while real-world experiments are conducted on an Intel Core i9-14900KF CPU with an NVIDIA RTX 4080 GPU.

\subsection{Simulation Experiments}
\label{sec:sim_experiments}

\begin{figure}[t]
    \centering
    \includegraphics[width=0.45\textwidth]{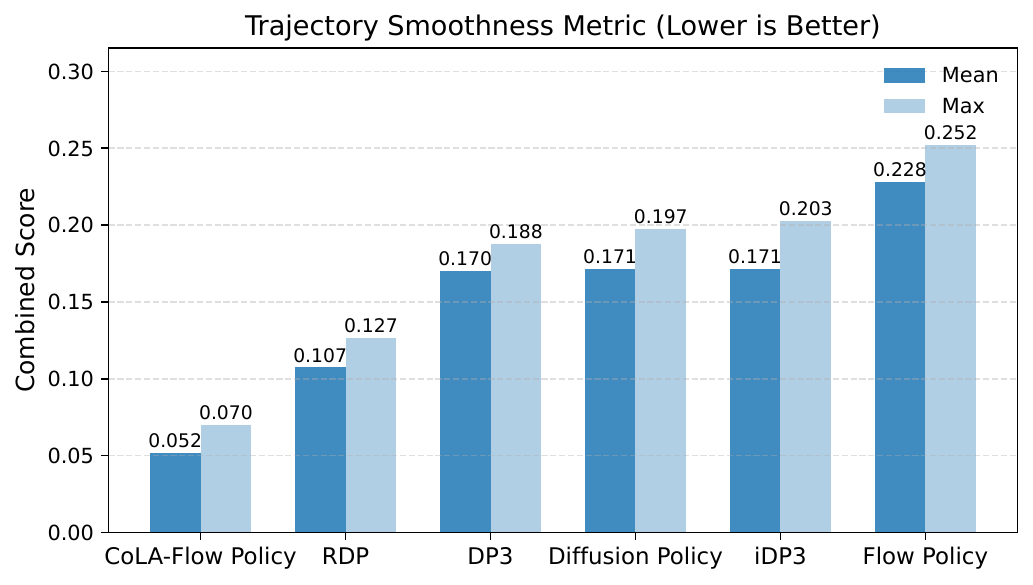}
    \caption{Trajectory smoothness across simulated tasks; lower is smoother.}
    \label{fig:sim_smoothness}
\end{figure}

Quantitative simulation results are summarized in Table~\ref{tab:sim_results} and Fig.~\ref{fig:sim_smoothness}, evaluating task success, inference latency, and trajectory smoothness.

\begin{table*}[t]
  \centering
    \caption{
    Simulation results. Success rates (\%) are mean $\pm$ standard deviation. Avg. Success averages Adroit and MetaWorld; $\Delta$ Success / $\Delta$ Time denote absolute differences from DP3 in average success and per-step latency.
    }
  \label{tab:sim_results}
  \renewcommand{\arraystretch}{1.15}
  \begin{tabular}{lcccccc}
    \toprule
    \textbf{Method} &
    \textbf{Success Rate in Adroit} &
    \textbf{Success Rate in MetaWorld} &
    \textbf{Avg. Success Rate} &
    $\boldsymbol{\Delta}$\textbf{ Success} &
    \textbf{Avg. Time} &
    $\boldsymbol{\Delta}$\textbf{ Time} \\
    \midrule
    Diffusion Policy &
    $26.0 \pm 4.0$ &
    $37.3 \pm 2.3$ &
    $31.7 \pm 6.9$ &
    $-34.0$ &
    $35.7 \text{ ms}$ &
    $-20.2$ \\
    Flow Policy &
    $56.7 \pm 2.3$ &
    $65.3 \pm 2.3$ &
    $61.0 \pm 5.2$ &
    $-4.7$ &
    $\mathbf{6.1 \text{ ms}}$ &
    $\mathbf{-49.8}$ \\
    DP3 &
    $62.0 \pm 2.0$ &
    $69.3 \pm 3.1$ &
    $65.7 \pm 4.6$ &
    $0.0$ &
    $55.9 \text{ ms}$ &
    $0.0$ \\
    iDP3 &
    $62.7 \pm 2.3$ &
    $71.3 \pm 2.3$ &
    $67.0 \pm 5.2$ &
    $+1.3$ &
    $58.9 \text{ ms}$ &
    $+3.0$ \\
    RDP &
    $70.0 \pm 2.0$ &
    $76.0 \pm 2.0$ &
    $73.0 \pm 3.7$ &
    $+7.3$ &
    $57.3 \text{ ms}$ &
    $+1.4$ \\
    \rowcolor{blue!10}
    \textbf{CoLA-Flow Policy (Ours)} &
    $\mathbf{76.0 \pm 2.0}$ &
    $\mathbf{80.7 \pm 3.1}$ &
    $\mathbf{78.3 \pm 3.4}$ &
    $\mathbf{+12.6}$ &
    $7.5 \text{ ms}$ &
    $-48.4$ \\
    \bottomrule
  \end{tabular}
\end{table*}

\paragraph{Trajectory Smoothness}
CoLA-Flow Policy achieves the lowest average smoothness score of $0.052$, reducing the metric by $51.4\%$, $69.4\%$, and $77.2\%$ relative to RDP, DP3, and Flow Policy, respectively.
RDP is smoother than other diffusion-based baselines, suggesting that latent action representations improve trajectory stability.
However, CoLA-Flow Policy further reduces high-frequency oscillations by modeling temporally coherent continuous latent trajectories.

\paragraph{Task Success Rate and Response Time}
Table~\ref{tab:sim_results} summarizes task success rates and inference latency across simulated benchmarks.
CoLA-Flow Policy achieves the highest average success rate of $78.3\%$, outperforming Flow Policy by $17.3$ percentage points, DP3 by $12.6$ percentage points, and RDP by $5.3$ percentage points.
Although Flow Policy has the lowest latency of $6.1$\,ms, CoLA-Flow Policy remains in the same real-time regime with $7.5$\,ms per control step.
Compared with DP3, CoLA-Flow Policy is approximately $7.5\times$ faster while achieving higher task success, showing a favorable success--latency trade-off.

Overall, simulation results show that CoLA-Flow Policy improves task success and trajectory smoothness while preserving the inference efficiency of flow-based policies.

\subsection{Real-World Experiments}
\label{sec:real_world_experiments}

\begin{figure}[t]
    \centering
    \includegraphics[width=0.45\textwidth]{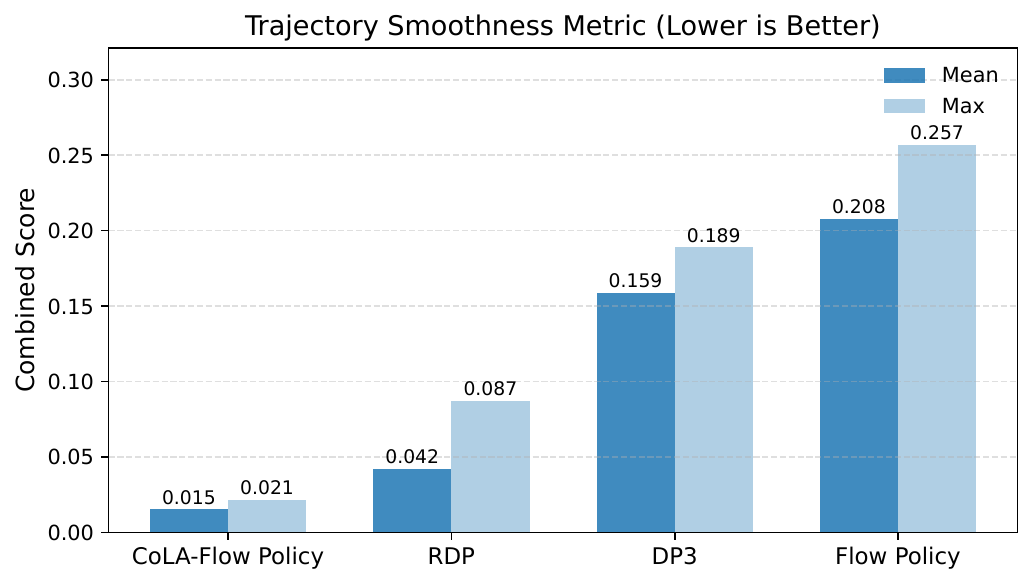}
    \caption{Trajectory smoothness across real-world tasks; lower is smoother.}
    \label{fig:realworld_smoothness}
\end{figure}

Table~\ref{tab:realworld_results} and Fig.~\ref{fig:realworld_smoothness} summarize real-world success, latency, and smoothness.

\begin{table*}[t]
  \centering
  \caption{
    Real-world robot results.
    Success rates (\%) are mean $\pm$ standard deviation over three 10-trial runs.
    Avg. Success averages all tasks; Avg. Response Time is per-step policy latency.
    }

  \label{tab:realworld_results}
  \renewcommand{\arraystretch}{1.15}
  \begin{tabular}{lcccccc}
    \toprule
    \textbf{Method} &
    \textbf{Pick \& Place} &
    \textbf{Pick \& Place (Gripper)} &
    \textbf{Peg-in-Hole} &
    \textbf{Obstacle Avoidance} &
    \textbf{Avg. Success} &
    \textbf{Avg. Response Time} \\
    \midrule
    DP3 &
    $66.7 \pm 5.8$ &
    $80.0 \pm 5.8$ &
    $35.0 \pm 5.8$ &
    $50.0 \pm 10.0$ &
    $57.9$ &
    $29.29 \text{ ms}$ \\
    RDP &
    $73.3 \pm 5.8$ &
    $93.3 \pm 5.8$ &
    $50.0 \pm 0.0$ &
    $63.3 \pm 5.8$ &
    $70.0$ &
    $31.82 \text{ ms}$ \\
    Flow Policy &
    $60.0 \pm 0.0$ &
    $73.3 \pm 5.8$ &
    $30.0 \pm 10.0$ &
    $46.7 \pm 5.8$ &
    $52.5$ &
    $\mathbf{6.54 \text{ ms}}$ \\
    \rowcolor{blue!10}
    \textbf{CoLA-Flow Policy} &
    $\mathbf{83.3 \pm 5.8}$ &
    $\mathbf{96.7 \pm 5.8}$ &
    $\mathbf{60.0 \pm 0.0}$ &
    $\mathbf{70.0 \pm 5.8}$ &
    $\mathbf{77.5}$ &
    $8.59 \text{ ms}$ \\
    \bottomrule
    \end{tabular}
\end{table*}

\paragraph{Trajectory Smoothness}
As shown in Fig.~\ref{fig:realworld_smoothness}, CoLA-Flow Policy achieves the lowest mean and maximum smoothness scores, indicating reduced jerk and high-frequency oscillations during real-world execution.
Compared with Flow Policy, the mean smoothness score is reduced by over $93.7\%$.
RDP also improves smoothness over action-space baselines, but remains less smooth than CoLA-Flow Policy, highlighting the benefit of continuous latent action modeling for stable execution.

\paragraph{Task Success Rate and Response Time}
Table~\ref{tab:realworld_results} reports real-world task success rates and inference latency.
CoLA-Flow Policy achieves the highest average success rate of $77.5\%$, outperforming DP3, RDP, and Flow Policy by $19.6$, $7.5$, and $25.0$ percentage points, respectively.
Although Flow Policy has the lowest latency of $6.54$\,ms, its unstable trajectories lead to lower task success.
CoLA-Flow Policy achieves substantially higher success with a response time of only $8.59$\,ms, and remains over $3\times$ faster than DP3 and RDP\@.
Overall, real-world results confirm that CoLA-Flow Policy provides a strong balance between inference efficiency, trajectory smoothness, and execution robustness.

\subsection{Long-Horizon and Latent Representation Ablation}
\label{sec:long_horizon_ablation}

To evaluate latent action representations in long-horizon manipulation, we compare Flow Policy, VQ-Flow (Flow Policy with a discrete VQ-VAE latent representation and MLP encoder), and CoLA-Flow Policy.
We define a long-horizon task as one whose execution duration exceeds 100\,s and involves multiple target objects.
The comparison is conducted on a multi-object pick-and-place task.

\begin{figure}[t]
    \centering
    \includegraphics[width=0.45\textwidth]{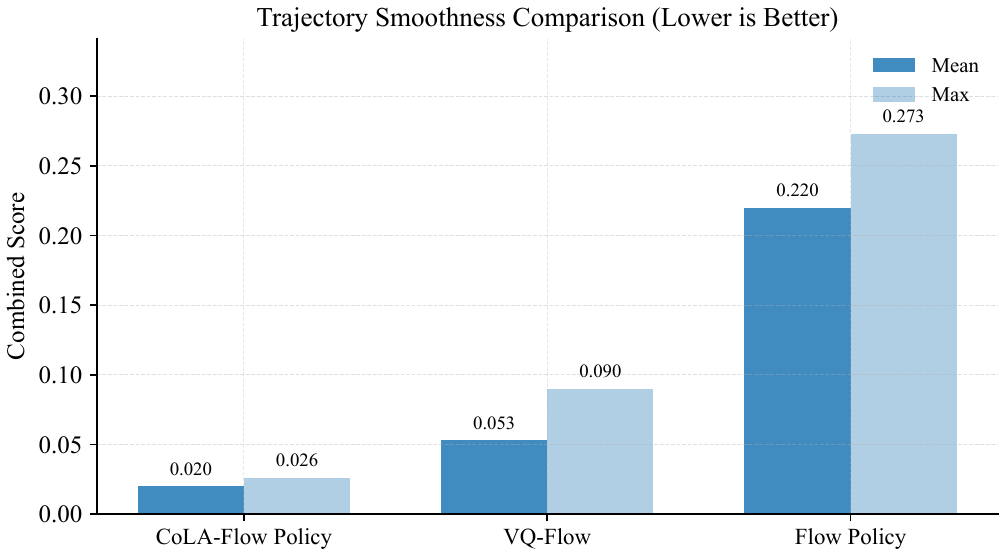}
    \caption{
    Trajectory smoothness comparison in the long-horizon multi-object pick-and-place task.
    Lower values indicate smoother execution.
    }
    \label{fig:abl_long_smooth}
\end{figure}

\begin{figure}[t]
    \centering
    \includegraphics[width=0.45\textwidth]{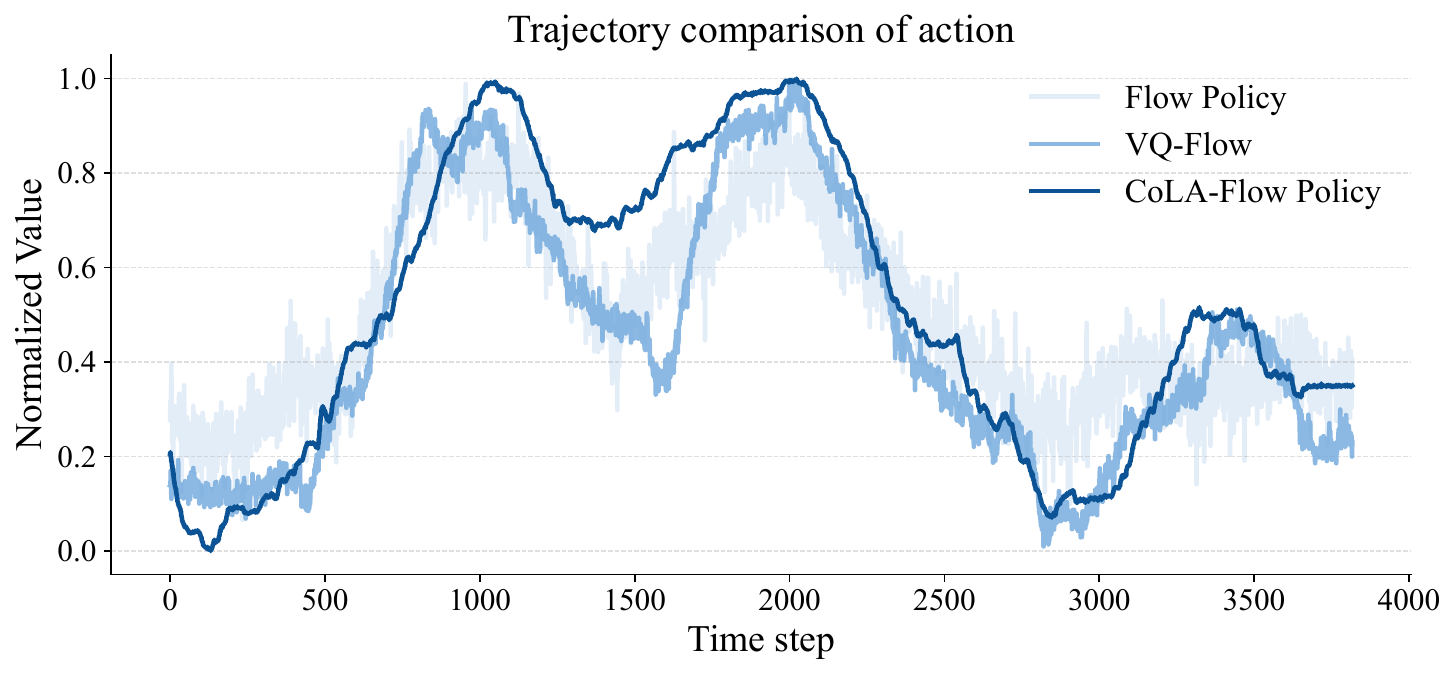}
    \caption{
    Long-horizon joint trajectories under identical task settings: Flow Policy exhibits high-frequency oscillations, VQ-Flow reduces abrupt variations, and CoLA-Flow Policy produces the smoothest and most temporally coherent trajectories.
    }
    \label{fig:abl_long_traj}
\end{figure}

\paragraph{Trajectory Smoothness and Visualization}
As shown in Fig.~\ref{fig:abl_long_smooth}, CoLA-Flow Policy achieves the lowest smoothness score of $0.02$, compared with $0.053$ for VQ-Flow and $0.22$ for Flow Policy.
This indicates that latent action modeling substantially reduces jerk and high-frequency oscillations, while the continuous latent representation further improves temporal consistency over the discrete latent variant.
Fig.~\ref{fig:abl_long_traj} further visualizes this effect: Flow Policy exhibits pronounced oscillations, VQ-Flow produces more stable trajectories, and CoLA-Flow Policy yields the smoothest joint motions.

\begin{table}[t]
  \centering
  \caption{Long-horizon performance under different action representations.}
  \label{tab:long_horizon_results}
  \renewcommand{\arraystretch}{1.15}
  \begin{tabular}{lcc}
    \toprule
    \textbf{Method} &
    \textbf{Success Rate} &
    \textbf{Response Time} \\
    \midrule
    Flow Policy &
    $20.0\%$ &
    $\mathbf{6.61 \text{ ms}}$ \\
    VQ-Flow &
    $70.0\%$ &
    $8.52 \text{ ms}$ \\
    \rowcolor{blue!10}
    \textbf{CoLA-Flow Policy} &
    $\mathbf{80.0\%}$ &
    $8.60 \text{ ms}$ \\
    \bottomrule
  \end{tabular}
\end{table}

\paragraph{Task Success Rate and Response Time}
Table~\ref{tab:long_horizon_results} shows that Flow Policy achieves only a $20.0\%$ success rate despite its lowest response time, mainly because severe oscillations often trigger protective stops.
VQ-Flow improves the success rate to $70.0\%$, while CoLA-Flow Policy further increases it to $80.0\%$ with nearly the same response time.
Thus, continuous latent action modeling improves long-horizon stability with minimal additional latency.

\subsection{Modality and Perturbation Ablation}
\label{sec:modality_perturb_ablation}

We further evaluate point-cloud encoding and wrist-camera conditioning in real-world execution, denoting the DP3 point-cloud encoder variant as \emph{CoLA-DP3Enc} and the variant without wrist-camera input as \emph{CoLA-NoWrist}.

\begin{figure}[t]
    \centering
    \includegraphics[width=0.45\textwidth]{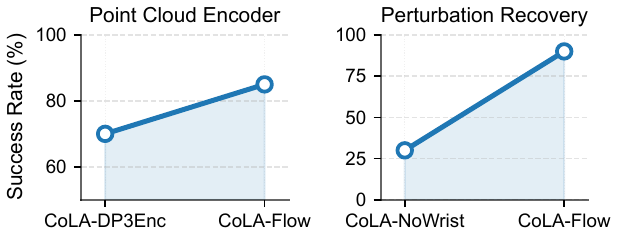}
    \caption{
    Modality and perturbation ablation.
    Left: CoLA-DP3Enc vs. CoLA-Flow Policy on real-world pick-and-place.
    Right: CoLA-NoWrist vs. CoLA-Flow Policy under small target perturbations.
    }
    \label{fig:abl_modality_perturb}
\end{figure}

As shown in Fig.~\ref{fig:abl_modality_perturb} (left), replacing our geometry-aware encoder with the DP3 encoder reduces the pick-and-place success rate from $85.0\%$ to $70.0\%$, indicating the benefit of the proposed point-cloud representation.

For wrist-camera conditioning, we apply target perturbations with displacement magnitudes of $30\%$--$60\%$ of the maximum end-effector opening range and evaluate task recovery.
As shown in Fig.~\ref{fig:abl_modality_perturb} (right), CoLA-Flow Policy achieves $90.0\%$ success, while CoLA-NoWrist drops to $30.0\%$.
The large drop of \emph{CoLA-NoWrist} shows that wrist-view local feedback is important for perturbation recovery. This also helps explain why RDP and CoLA-Flow Policy, which both use wrist-view feedback in real-world experiments, are the two strongest real-world methods; CoLA-Flow Policy's remaining advantage further reflects the benefit of continuous latent flow and geometry-aware conditioning.

\section{CONCLUSION}

This work addresses real-time robotic manipulation in long-horizon, high-dimensional control, where generative policies must balance expressiveness, latency, and stability.
We propose \textbf{CoLA-Flow Policy}, which performs flow matching in a coherent latent action space to generate temporally consistent trajectories with near-single-step inference.
Its continuous latent representation decouples global motion structure from low-level control noise, while geometry-aware point-cloud conditioning and execution-time decoding modulation incorporate scene and wrist-camera feedback without perturbing latent dynamics.
Simulation, real-robot, long-horizon, modality, and perturbation ablations validate the roles of continuous latent actions, geometry-aware encoding, and wrist-camera modulation, yielding smoother trajectories, higher success rates, and faster inference than diffusion-based, discrete-latent, and raw action-space flow baselines.

The current system still uses lightweight multimodal fusion and does not explicitly model complex contact dynamics.
Future work will explore richer multimodal representations and contact-aware latent dynamics for more challenging contact-rich manipulation.

\bibliographystyle{IEEEtran}
\bibliography{references}

\vfill
\end{document}